\documentclass[12pt,a4paper, oneside]{article}
\usepackage[utf8]{inputenc}
\usepackage{setspace}
\usepackage[letterpaper, right=2.54cm, left=2.54cm, top=2.54cm, bottom=2.54cm]{geometry}
\usepackage{natbib}
\usepackage{enumitem}
\usepackage{amsfonts}
\usepackage{color}
\definecolor{CardinalRed}{cmyk}{0,1,0.65,0.34} 
\usepackage[colorlinks,linkcolor=CardinalRed,citecolor=CardinalRed,urlcolor = CardinalRed]{hyperref}
\usepackage{dirtytalk}
\usepackage{graphicx}
\usepackage[flushmargin, bottom]{footmisc}
\usepackage[group-separator={,}]{siunitx}
\usepackage{tabularx}
\newcolumntype{Y}{>{\centering\arraybackslash}X}
\usepackage{booktabs}
\newcommand{\blue}{\textcolor{blue}}
\usepackage{float}
\usepackage{comment}

\title{Benchmark Illusion: Disagreement among LLMs and Its Scientific Consequences}

\author{Eddie Yang\thanks{Department of Political Science, Purdue University, \blue{eddieyang@purdue.edu}.} \and Dashun Wang\thanks{Kellogg School of Management, Center for Science of Science and Innovation, Northwestern Innovation Institute, Ryan Institute on Complexity, and McCormick School of Engineering, Northwestern University, \blue{dashun.wang@northwestern.edu}}}

\date{}

\begin{document}
\maketitle

\begin{abstract}

Benchmarks underpin how progress in large language models (LLMs) is measured and trusted. Yet our analyses reveal that apparent convergence in benchmark accuracy can conceal deep epistemic divergence. Using two major reasoning benchmarks -- MMLU-Pro and GPQA -- we show that LLMs achieving comparable accuracy still disagree on 16-66\% of items, and 16-38\% among top-performing frontier models. These discrepancies suggest distinct error profiles for different LLMs. When such models are used for scientific data annotation and inference, their hidden disagreements propagate into research results: in re-analyses of published studies in education and political science, switching the annotation model can change estimated treatment effects by more than 80\%, and in some cases reverses their sign. Together, these findings illustrate a benchmark illusion, where equal accuracy may conceal disagreement, with model choice becoming a hidden yet consequential variable for scientific reproducibility.

\end{abstract}

\thispagestyle{empty}
\clearpage
\newpage
\setcounter{page}{1}
\onehalfspacing

\section{Introduction}

Large language models (LLMs) have increasingly become embedded in scientific practice \citep{gao2024quantifying, gil2025accelerating, zhang2025exploring}. Researchers now use them to analyze texts, extract variables from documents, summarize evidence, and even simulate survey respondents \citep{agrawal2022large, gilardi2023chatgpt, argyle2023out, cui2025large}. This appeal largely rests on two related premises. First, LLMs demonstrate impressive performance on standard benchmarks for reasoning and knowledge. Second, once a model has \say{cleared} these benchmarks, it is often treated as a plug-and-play component: a high-performing tool that can be integrated into scientific workflows without fundamentally changing how we reason about inference.

Both premises depend on benchmarks. Common benchmarks such as MMLU-Pro \citep{wang2024mmlu} and GPQA \citep{rein2024gpqa} have become the de facto currency for judging LLM capabilities. As scores on these benchmarks have risen, it has become common, both implicitly and explicitly, to treat benchmark accuracy as a proxy for capability, reliability, and even interchangeability. If two models achieve similar scores, the working assumption is that they possess similar levels of knowledge and reasoning capabilities, and that they can be substituted for one another in downstream applications.

In this paper, we show that this assumption -- which we call the benchmark illusion -- often fails, because models with similar benchmark accuracy can diverge in ways that affect scientific inference.\footnote{See also \citet{d2022underspecification} regarding \textit{underspecification}, the phenomenon where equivalent test-set accuracy masks substantive differences in model behavior.} We begin by examining the extent to which ostensibly comparable models actually agree when evaluated on identical items. Using model predictions from \citet{wang2024mmlu} on MMLU-Pro and the HELM benchmark suite \citep{liang2022holistic} on GPQA, we find substantial pairwise disagreement. Among models with comparable accuracy, disagreement ranges from 16–66\% on MMLU-Pro and 17–65\% on GPQA. Crucially, this divergence persists even among top-performing frontier models with accuracy above 60\%, where disagreement remains between 16\% and 38\%. Because prompts and decoding are held fixed, these discrepancies reflect systematic differences in how models represent and reason about information. In effect, models that each \say{get 7 out of 10 questions right} systematically differ in \emph{which} 7 they get right.

For many applications, this distinction may be secondary. For science, it is central. When LLMs are used to construct outcomes or covariates from unstructured data, disagreement on which items are misclassified can translate directly into differences in estimated effects, subgroup patterns, or even the sign of a result \citep{egami2024using, baumann2025large}. In other words, for the scientific use of AI, what matters is not only how often a model is wrong, but where and for whom it is wrong. To make this precise, we use a simple measurement error framework to show how systematic prediction errors can bias estimates in regression settings, even when overall accuracy is high. Building on this framework, we run a simulation in which three annotators, including one relatively inaccurate but with random errors and two with very high accuracy but systematic errors, produce markedly different treatment effect estimates. The two high accuracy annotators, despite similar aggregate performance, induce larger biases than the less accurate annotator and in opposite directions.

We then examine how these mechanisms play out in practice by reanalyzing two published studies. The first is a large-scale randomized trial of a literacy intervention in education \citep{kim2021improving}, in which human graders scored student essays using a detailed rubric. Replacing the original scores with those from eight high-performing LLMs, we find that while all models recover a positive treatment effect, the estimated magnitude varies by roughly 80\%, ranging from 0.19 to 0.35 against an original estimate of 0.44. The second case study, from political science, examines selective attribution in Russian state media \citep{rozenas2019autocrats}. Here, the consequences of model choice are even more severe. While some models replicate the original finding that officials are more likely to be credited with good news, others suggest the opposite, implying officials are more likely to be blamed for bad news. In both case studies, model choice alone, while holding data, design, and code fixed, changes the quantitative conclusions in substantively meaningful ways.

Our findings speak to, and build upon, a rapidly growing literature on LLMs as tools for scientific research.\footnote{See e.g., \citealt{hao2024ai, kusumegi2025scientific, westwood2025potential} for recent work in this area.} In computational social science, \citet{ziems2024can} show that LLMs can act as zero-shot annotators, arguing that they can meaningfully augment the research pipeline. A separate line of work in statistical methodology cautions that using predicted labels as true outcomes can bias estimates, and proposes correction methods for valid inference \citep{angelopoulos2023prediction, egami2024using}. Recent work further documents how seemingly small and defensible changes in model, prompt, or configuration can lead to \say{LLM hacking,} flipping statistical conclusions from analyses of various annotated variables \citep{reiss2023testing, baumann2025large, carlson2025use}. Meanwhile, scholars also highlight the volatility of proprietary LLM APIs \citep{barrie2024replication}, while others argue that non-state-of-the-art models remain useful if errors and result sensitivity are properly characterized \citep{bisbee2025human}.

We extend this work in three ways. First, we shift attention upstream from particular annotation tasks to the structure of disagreement among models themselves on canonical, gold-standard benchmarks. Rather than asking whether a given model agrees with human labels, we ask how much high-performing models agree with one another, and show that benchmark accuracy can mask substantial epistemic variation. Second, we show how this variation translates into materially different scientific conclusions, even when researchers restrict themselves to models that look equivalent by conventional metrics. In this sense, our results make concrete a new dimension of researcher degrees of freedom: the choice of LLM becomes analogous to a specification choice, with similar potential to introduce forking paths \citep{gelman2013garden}. Third, we use these findings to argue that current evaluation regimes are misaligned with scientific needs. For many scientific applications, what matters is not only how accurate a model is on average, but which errors it makes, how those errors relate to variables of interest, and how stable its behavior is across replications and over time.

Together, these contributions highlight that LLMs introduce a distinctive form of variation: systematic disagreement among automated research agents that are, by conventional standards, all \say{good} models. Just as scientists have developed frameworks to account for disagreements among human coders and researchers \citep{krippendorff2018content}, we must now confront the unique nature of disagreements among LLMs to ensure the continued credibility of science in an age of artificial intelligence.

\section{LLMs Make Different Mistakes}

We first test a core premise of the benchmark illusion: that models achieving similar accuracy scores are functionally equivalent. To do this, we examine answers by different models on two authoritative reasoning benchmarks: MMLU-Pro, a broad dataset of 12,032 multiple-choice questions across 14 disciplines, and GPQA, a challenging dataset of 448 expert-level questions in biology, physics, and chemistry. Both benchmarks are designed to test the knowledge and reasoning abilities of LLMs.

\begin{figure}[h]
    \centering
    \caption{MMLU-pro -- Pairwise Model Proportion of Disagreement}\label{fig:mmlupro}
    \includegraphics[scale=0.42]{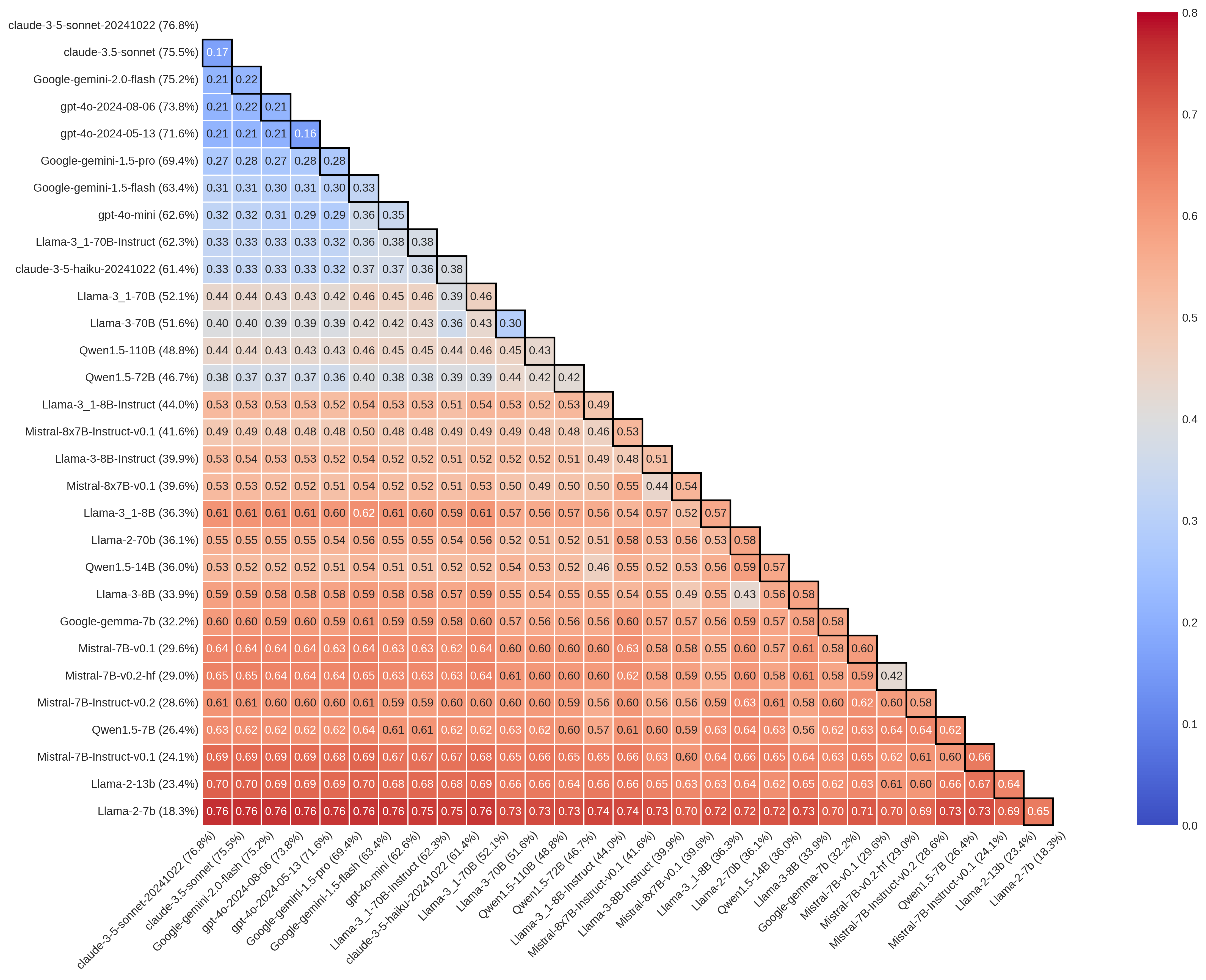}
\end{figure}

\begin{figure}[h]
    \centering
    \caption{GPQA -- Pairwise Model Proportion of Disagreement}\label{fig:gpqa}
    \includegraphics[scale=0.42]{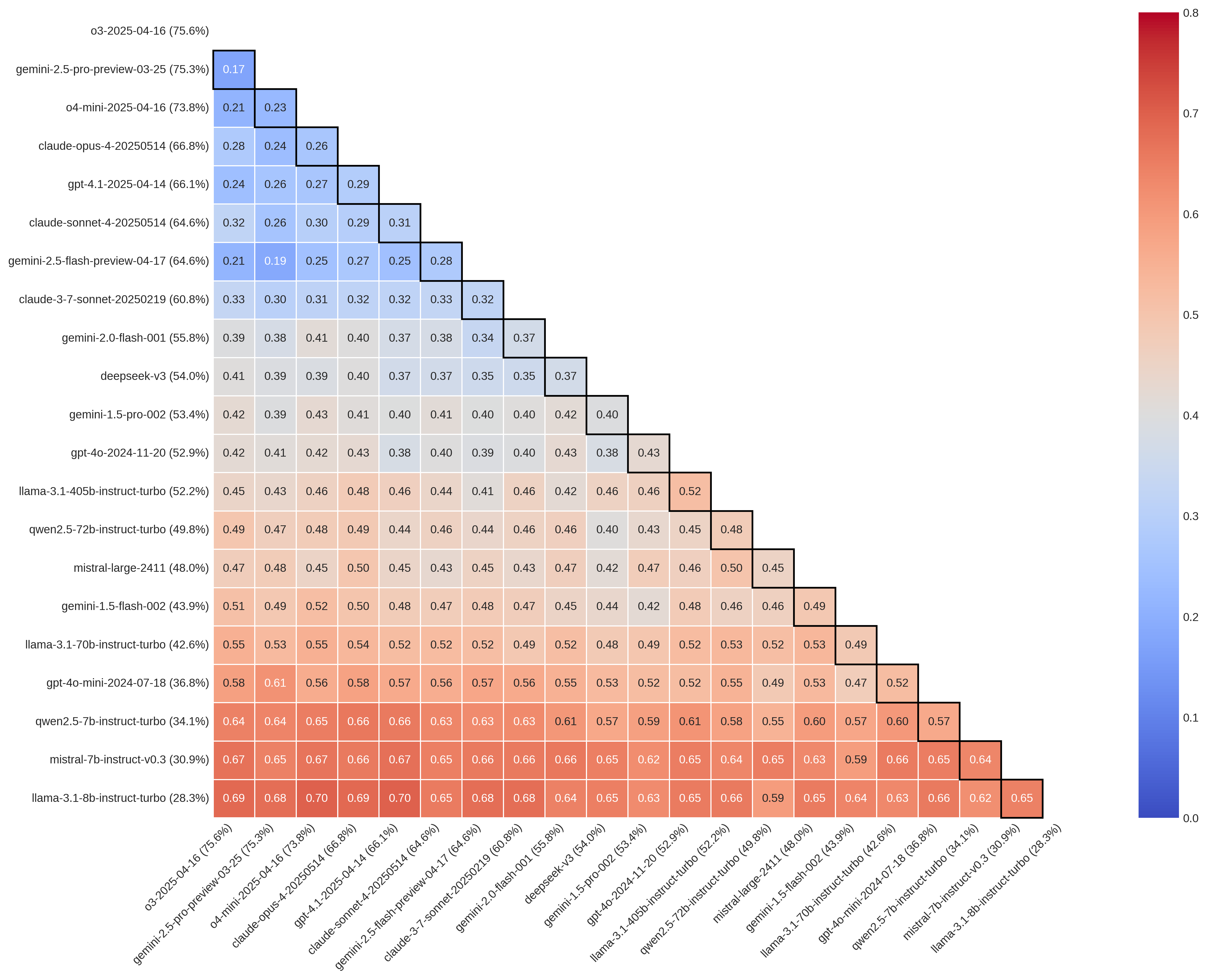}
\end{figure}

Drawing on model predictions from the creators of MMLU-Pro and the Holistic Evaluation of Language Models (HELM) benchmark suite, we calculate the proportion of questions on which any two models provide different answers. The results, visualized in Figures \ref{fig:mmlupro} and \ref{fig:gpqa}, reveal a striking degree of divergence. Across all models, the pairwise disagreement rate is substantial, ranging from 16\% to 76\% on MMLU-Pro and 17\% to 70\% on GPQA.

Crucially, this disagreement persists even among models with comparable overall accuracy. As highlighted by the black borders in the figures, even models with the closest accuracy scores exhibit substantial divergence. For the top-tier models (accuracy $> 60\%$), disagreement rates still range from 16\% to 38\% on MMLU-Pro and 17\% to 32\% on GPQA. Because these evaluations use fixed prompts and deterministic decoding, this variation is not random noise. Instead, it points to systematic differences in how models represent and reason about information. In essence, models with similar benchmark accuracies can arrive at those scores via fundamentally different trajectories, answering different subsets of questions correctly. This finding challenges the assumption that accuracy scores on benchmarks capture a complete profile of a model's underlying capabilities.

\section{The Dangers of LLM Difference for Scientific Analysis}
The persistent and widespread nature of LLM disagreement, even among the frontier models, suggests that different LLMs may have distinct systematic biases and failure modes. For a researcher using an LLM to annotate data, the choice of model is therefore not a neutral one. It is an implicit choice of a particular error profile. If this error profile systematically correlates with key variables in a study -- for instance, if one model is more prone to misclassifying texts from a specific demographic or treatment group -- the resulting annotations will be biased in a structured way. This, as we will show, can fundamentally alter the results of downstream statistical analyses. 

To illustrate the problem of prediction errors on downstream statistical analyses, consider a simple case where the outcome variable $Y$ is a latent characteristics of unstructured data that requires annotation, and the researcher is interested in its relationship with independent variables $ \mathbf{X} = \{X_{1}, X_{2}\}$ given by the linear model,
$$\mathbb{E}[Y \mid \mathbf{X}] = \mathbf{X}^{T}\mathbf{\beta}$$
To make things concrete, suppose a clinical researcher is investigating the safety profile of a new antibiotic drug. In this case, $Y$ is a binary indicator for whether a patient’s unstructured electronic health record (EHR) notes mention a specific adverse event (e.g., rash or allergic reaction). $X_{1}$ is an indicator for whether the patient received the new drug versus a standard comparator, and $X_{2}$ is a continuous variable representing the number of days the patient was hospitalized.

The researcher does not observe the true $Y$ (manual review by a safety specialist), but instead uses an LLM to generate an annotated variable $\hat{Y}$. Using $\hat{Y}$, the researcher obtain the ordinary least squares estimates of $\mathbf{\beta}$, denoted as $\hat{\mathbf{\beta}}$. However, the estimates are unbiased only if prediction errors, $e = \hat{Y} - Y$, are zero on average across all different combinations of $\mathbf{X}$, i.e., 
$\mathbb{E}[e \mid \mathbf{X}] = 0$. In other words, for $\mathbf{\beta}$ to be unbiased, prediction errors must be uncorrelated with the independent variables, the true outcome variable, or any unobserved confounders \citep{egami2024using}. If $e$ is affected by either $\mathbf{X}$, $Y$, or an unobserved confounder, $U$, then $\mathbb{E}[e \mid \mathbf{X}] \neq 0$ and $\hat{\mathbf{\beta}}$ will be biased. In the clinical example, this means the LLM's error rate in detecting the adverse event must be the same for patients who truly experienced the event and those who did not ($Y$), for patients in the treatment and comparator groups ($X_{1}$), and across hospitalization time $X_{2}$.

Now imagine if the researcher has access to a number of different LLMs, denoted by $L_{1}$, $L_{2}$, ..., $L_{n}$. Since LLMs disagree, the error profile of each LLM will generally be different, i.e., $e_{L_{1}} \neq e_{L_{2}} \neq ...\neq e_{L_{n}}$. This means the magnitude and direction of the bias are potentially different across LLMs as well, i.e., $\mathbb{E}[\hat{\mathbf{\beta}}_{L_{1}} - \mathbf{\beta}] \neq \mathbb{E}[\hat{\mathbf{\beta}}_{L_{2}} - \mathbf{\beta}] \neq ...\neq \mathbb{E}[\hat{\mathbf{\beta}}_{L_{n}} - \mathbf{\beta}]$. As a result, not only do we get biased estimates, but the estimates will also depend on the choice of LLM, potentially making the findings less replicable.

\section{Implications for Science: Simulation}
We first use a simulation to clarify the causal mechanism by which LLM disagreements affect scientific conclusions. Specifically, we demonstrate two points: first, an annotator with excellent aggregate performance can still induce substantial bias if its errors are systematically correlated with variables of interest; and second, annotators with similar aggregate performance can induce very different biases.

\subsection{Simulation Design}
We simulate a dataset of N = 10,000 observations mirroring the antibiotic drug research scenario described in the previous section. The data generating process follows a logistic regression model, where a binary outcome $Y$ is determined by a binary treatment variable $X_{1} \sim Bernoulli(0.5)$ and a continuous control variable $X_{2}  \sim N(0, 1)$. The true model is: $$logit(P(Y=1)) = -1 + X_{1} + X_{2}$$

Because expert assessment of 10,000 EHR notes is slow and expensive, the clinical researcher may decide to use cutting-edge AI models to analyze the EHR notes instead. In particular, we examine the downstream consequences of three \say{AI annotators}.

\textbf{Annotator 1 (Low performance, random error):} As a baseline, Annotator 1 makes annotation mistakes completely at random, flipping any given label with a 15\% probability. This annotator is intentionally designed to be less performant on aggregate metrics, with an overall accuracy of 85.0\%.

\textbf{Annotator 2 (High performance, systematic error):} This annotator mimics a plausible LLM failure mode. It is highly accurate overall but possesses a subtle, systematic bias: it is more likely to misclassify a true positive case (Y=1) as a negative when that case belongs to the treatment group ($X_{1}=1$). This error occurs 25\% of the time for treated positive cases but only 1\% of the time for non-treated positive cases. This results in an annotator with a very high overall empirical accuracy of 93.6\%.

\textbf{Annotator 3 (High performance, systematic error):} Similar to Annotator 2, Annotator 3 is highly accurate but flips the bias: it misclassifies a true positive case (Y=1) as a negative only 3\% of the time for treated positive cases but 30\% of the time for non-treated positive cases. Annotator 3 has an empirical accuracy of 93.8\%.

\subsection{Results}
\begin{figure}[h]
\begin{center}
\caption{Simulation result}
\label{fig:sim}
\includegraphics[scale=0.54]{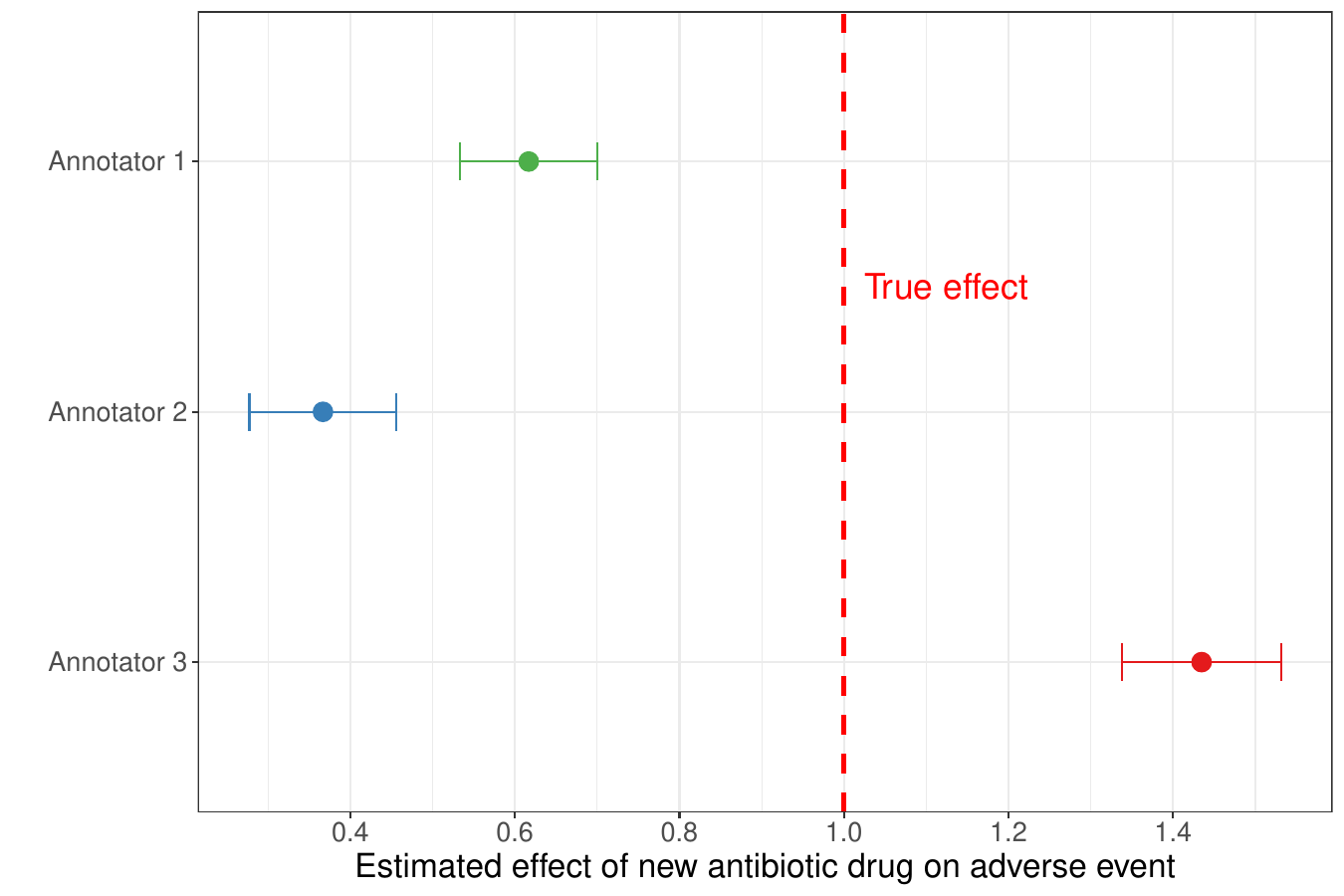}
\end{center}
\vspace{-2ex}
{\footnotesize \textbf{Notes:} The plot shows the estimated treatment effect from a logistic regression using data annotated by three simulated annotators. Annotator 1, though less performant on aggregate metrics, exhibits classical attenuation bias, which is less severe in this case. Annotators 2 and 3, despite their similar high performance, result in larger biases that are in the opposite direction.}
\end{figure}

We run three logistic regressions to estimate the effect of $X_{1}$, using annotations from each of the annotators as the observed outcome respectively. The results are presented in Figure \ref{fig:sim}.

The figure presents the estimated coefficients for the treatment variable ($X_{1}$) from the three models, with the true effect of $1.0$ shown as a red dashed line. For Annotator 1, whose errors are random, the estimated effect is approximately 0.62. This result shows the classic attenuation bias, where the estimate is pulled toward the null value of zero.

The results for Annotators 2 and 3, which have higher overall accuracy but systematic errors, are more striking. Annotator 2, which undercounts adverse events specifically in the treatment group, leads to a severe underestimation of the treatment effect (estimated coefficient = 0.37). This systematic error makes the drug appear safer than it truly is. In contrast, Annotator 3, which undercounts adverse events in the control group, produces a substantial overestimation of the effect (estimated coefficient = 1.43). By artificially lowering the number of adverse events of the comparator drug, it makes the new antibiotic drug seem more prone to triggering the adverse event.

The simulation provides a demonstration that, for statistical inference in research, the structure of an annotator's errors can be more consequential than its overall error rate. When errors correlate with variables of interest, state-of-the-art accuracy offers little protection against severe and misleading biases, a point we now explore with real-world data.

\section{Implications for Science: Empirical Applications}

To demonstrate the implications of LLM disagreement in real-world scientific studies, we reanalyze data from two studies, taken from education and political science respectively. For each of the studies, the outcome variable was annotated from text by either expert coders or crowdsource workers. We replicate the annotation process using different open-source and commercial LLMs. We then use the annotations by different LLMs in place of the original annotations and re-run the analyses of the original studies. Finally, we compare the estimates both among the different LLMs and with the original estimates.

\subsection{Case Study 1: Education}

Our first case study revisits a large-scale randomized controlled trial by \citet{kim2021improving}, which evaluated the effectiveness of a content literacy intervention called MORE (Model of Reading Engagement). The study involved 5,494 first- and second-grade students in 30 U.S. elementary schools. Students in the treatment group received thematic lessons designed to build domain knowledge and vocabulary in science and social studies, with the goal of improving their argumentative writing skills. In the original study, argumentative writing samples produced by students were scored by trained human raters using a detailed rubric that assessed the quality of the claim, evidence, and conclusion. The authors found that the intervention had a positive and statistically significant effect on students' argumentative writing, with an estimated effect size of 0.44 for social studies-related writing.

We obtained the student writing samples and replicated the original analysis. Instead of using the human-generated scores, we tasked eight different LLMs with scoring each essay according to the rubric described in the original paper. We then used the scores from each LLM as the outcome variable and re-estimated the treatment effect of the MORE intervention, using the same hierarchical linear model specification as the original study.

\begin{figure}[h]
\begin{center}
\caption{Reanalysis of \citet{kim2021improving}}
\label{fig:education}
\includegraphics[scale=0.5]{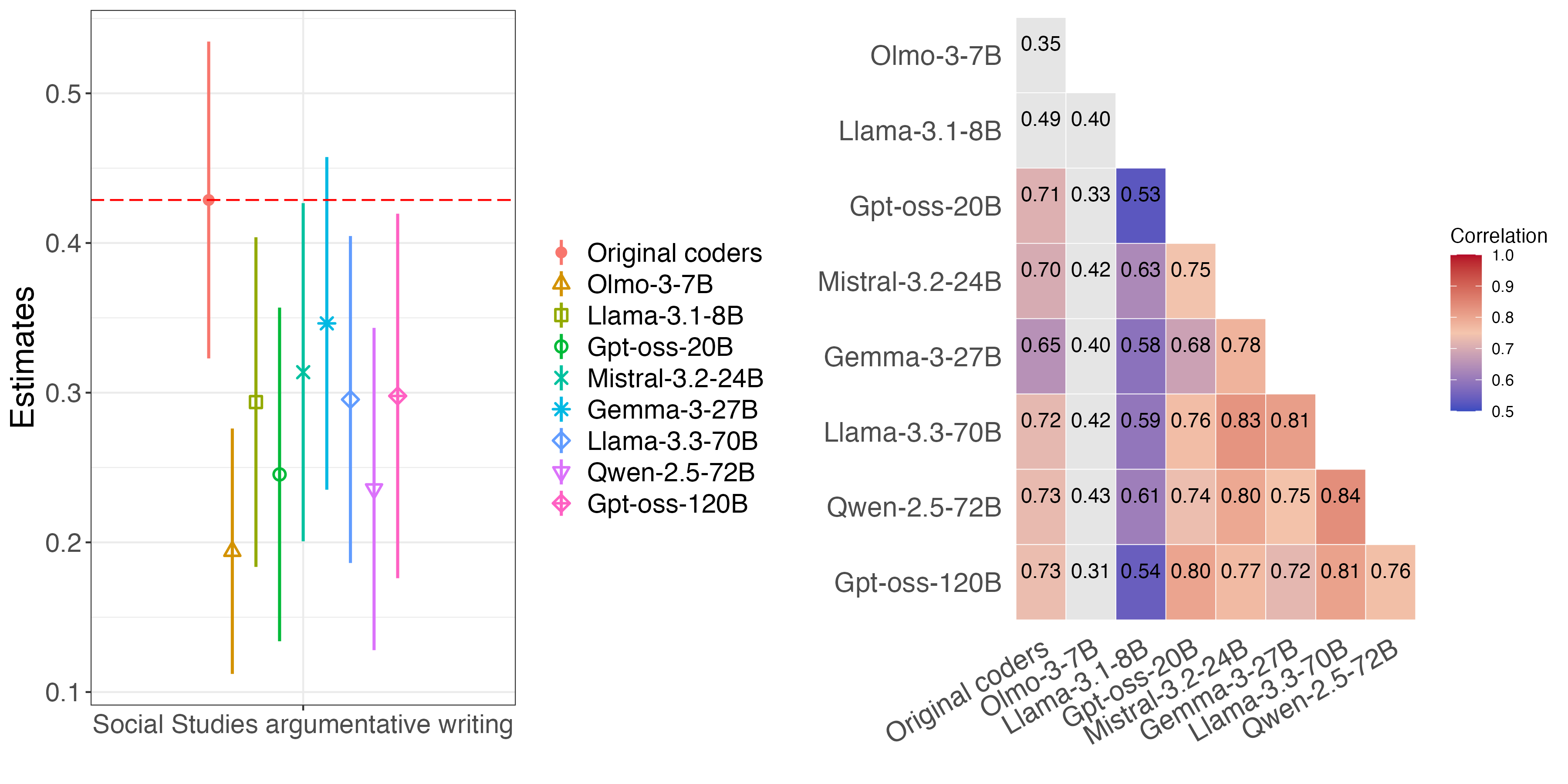}
\end{center}
\vspace{-2ex}
{\footnotesize \textbf{Notes:} The left panel shows the estimated treatment effect of a literacy intervention on student argumentative writing scores. Each point represents an estimate using scores from a different annotator (LLMs or the original human-coded study data), with bars indicating 95\% confidence intervals. The red dashed line marks the original study's point estimate. The right panel shows the correlation of the scores by a given pair of coders.}
\end{figure}

Figure \ref{fig:education} presents our reanalysis results. The left panel displays the estimated treatment effect on social studies argumentative writing for the original study (based on human coders) and for each of the eight LLMs. The right panel shows the correlations of the scores by a given pair of coders.

Our reanalysis reveals two key findings. First, while all LLMs correctly identify a positive and statistically significant treatment effect (all confidence intervals are above zero), the magnitude of this effect varies substantially depending on the model used for annotation. The estimated effects range from a low of 0.19 for Olmo-3-7B to a high of 0.35 for Gemma-3-27B. This represents an 84\% difference in the estimated impact of the educational intervention. This variation demonstrates that the choice of LLM can act as a powerful researcher degree of freedom, capable of substantially altering the quantitative conclusions of a study.

Second, the right panel illustrates that aggregate measures like correlation with a gold standard do not reliably predict the direction and magnitude of bias in downstream estimates. For instance, Gpt-oss-120B has the highest correlation with the original human scores (0.73) but produces a rather modest treatment effect estimate. Conversely, Gemma-3-27B, which yields the largest effect estimate among the LLMs, has a more moderate correlation of 0.65. This disconnect occurs because correlation is an aggregate measure that can conceal systematic biases. A model might agree well with human raters on average but still systematically under-score essays from the treatment group, for example, which would directly and artificially reduce the estimated treatment effect. This finding empirically illustrates the point from our simulation: the structure of a model’s errors, not just its overall agreement with a gold standard, is the critical factor for the validity of scientific conclusions.

\subsection{Case Study 2: Political Science}
Our second case study examines the work of \citet{rozenas2019autocrats}, who investigate how autocratic governments manipulate economic news. The authors challenge the conventional wisdom that autocrats simply censor bad news. Instead, they propose a strategy of \say{selective attribution}: state-controlled media systematically blame external factors (e.g., foreign governments, global markets) for bad economic news while attributing good economic news to the competence of domestic politicians. Using a corpus of daily news reports from Russia's largest state-owned television network, the original study employed crowdsourced human coders to classify news fragments. Their analysis confirmed their hypothesis, finding that Russian officials were significantly more likely to be mentioned in the context of good economic news than bad economic news.

We reanalyze the news fragments from the original study. The key task for the annotators is to determine whether a given news was attributed to \say{Russian officials.} We tasked eight different LLMs with this classification task for the entire corpus. We then used the labels generated by each LLM to re-estimate the \say{relative attribution} to Russian officials - the difference in the predicted probability of attributing Russian officials in the context of good news versus bad news. This is the central quantity of interest in the original paper.

\begin{figure}[h]
\begin{center}
\caption{Reanalysis of \citet{rozenas2019autocrats}}
\label{fig:polsci}
\includegraphics[scale=0.5]{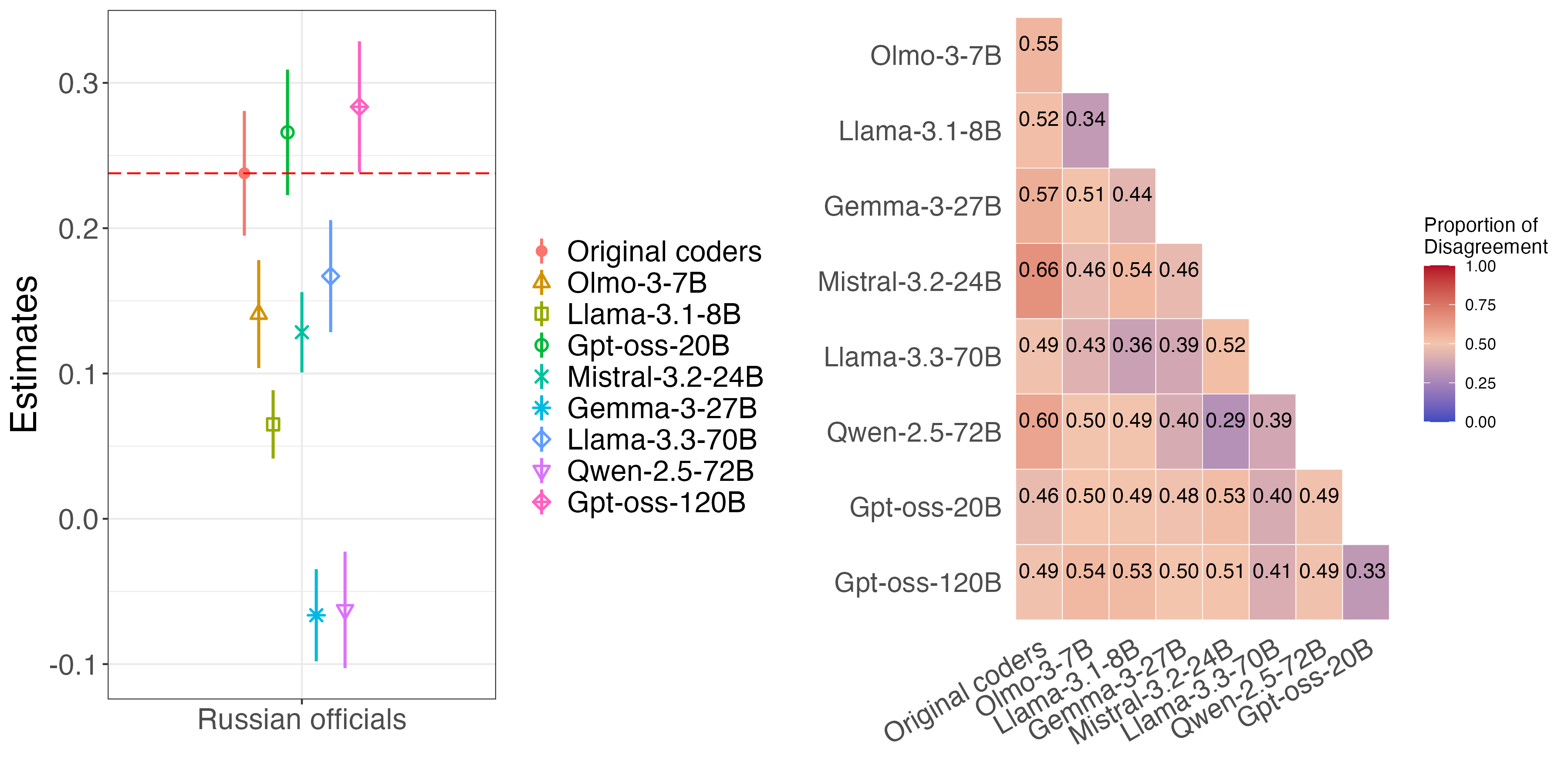}
\end{center}
\vspace{-2ex}
{\footnotesize \textbf{Notes:} The left panel shows the estimated relative attribution of economic news to \say{Russian officials.} Each point is an estimate using annotations from a different source (LLMs or the original crowdsourced data), with bars indicating 95\% confidence intervals. The red dashed line marks the original study's point estimate. A positive value indicates that good news is more likely to be attributed to Russian officials than bad news. The right panel shows the pairwise proportion of disagreement among the different annotators.}
\end{figure}

Figure \ref{fig:polsci} shows the results of our reanalysis. The left panel plots the estimated effect for the original study and for each of the eight LLMs. The right panel displays the pairwise proportion of disagreement among the different annotators.

The results present a stark illustration of how the choice of an LLM can fundamentally alter the conclusions of a scientific study. The original study (red) found a positive and statistically significant effect, with an estimate of approximately 0.24. This supports the theory of selective attribution. Our reanalysis shows that while some LLMs (like Gpt-oss-20B and Gpt-oss-120B) replicate this positive and significant finding, a number of LLMs do not. Two of the eight models we tested (Gemma-3-27B and Qwen-2.5-72B) produce estimates that reach the opposite conclusion - that Russian officials are more likely to be blamed for bad news than to be attributed with good news. 

The right panel of Figure \ref{fig:polsci} reveals the source of this instability: high levels of pairwise disagreement among the annotators. The proportion of disagreement is substantial, ranging from 29\% to as high as 66\%. Importantly, even state-of-the-art models disagree substantially both with the original crowdsourced labels and among themselves. For example, Gemma-3-27B and Gpt-oss-120B, which produce estimates on opposite sides of zero, disagree on 50\% of the news fragments they classify. This demonstrates that the models are not just exhibiting random error but are applying systematically different interpretations to the annotation task. For a task as nuanced as identifying implicit attribution in political news, these distinct \say{world models} lead to fundamentally different datasets and, consequently, opposing scientific conclusions. This case illustrates that when LLMs are used for scientific inference, the choice of model can be as consequential as the choice of statistical method or experimental design, with the potential to determine the outcome of the research itself.

\section{Discussion}

Large language models offer remarkable possibilities for scientific research. They can analyze complex texts, construct variables from unstructured data, help design and pretest instruments, and surface empirical patterns that would be prohibitively costly to uncover with human labor alone. Much of the current enthusiasm across the social, behavioral, and biomedical sciences reflects this potential, and strong benchmark performance has played a central role in legitimizing LLMs as reliable partners in research \citep{messeri2024artificial}.

Our analyses suggest that this comfort rests in part on a benchmark illusion. When benchmark scores converge, it is tempting to infer that models share a common underlying representation of knowledge and reasoning and are therefore interchangeable. We show that this inference can be fragile. On two widely used reasoning benchmarks, MMLU-Pro and GPQA, models that appear comparable by overall accuracy disagree on a substantial fraction of individual items. In downstream applications, these distinct error profiles carry real consequences. In our empirical reanalyses, swapping one high-performing model for another changes estimated effects by more than 80\% in the education case and reverses the direction of the key relationship in the Russian news case. Model identity thus becomes an additional design choice with substantive implications, rather than a purely technical implementation detail.

These observations force a reconsideration of model alignment. To date, model development has understandably focused on aligning LLMs to broad human preferences -- ensuring systems are helpful, harmless, and honest. While essential for safety and usability, this form of alignment is only indirectly related to the properties required for scientific measurement. Scientific applications demand replicability across time and ostensibly similar models, calibrated and interpretable uncertainty, and error patterns that do not systematically distort relationships among treatments, covariates, and outcomes. Our results indicate that models optimized for common benchmarks do not necessarily satisfy these criteria. Indeed, the very behaviors that make models agreeable to human users may introduce subtle regularities that interact with variables of interest in ways standard evaluations fail to reveal. 

This leads us to reconsider how LLM progress is evaluated and how models are developed for scientific use. Standard leaderboards primarily reward aggregate correctness on fixed benchmark sets. For many research workflows, however, it is important to understand not just whether a model is correct on average, but whether its behavior is stable under small perturbations, whether different high-performing models converge on the same answers, how uncertainty is calibrated, and whether errors are structured in ways that correlate with treatments, subgroups, or other design variables. The benchmark illusion we document therefore points to a mismatch between what current evaluations optimize and what \say{science-ready} models would need to guarantee. Our results are not only a caution for users of LLMs in science; they also motivate a complementary evaluation agenda that treats agreement, stability, and inferential robustness as first-class targets alongside accuracy. More broadly, this suggests a notion of scientific alignment: models evaluated and optimized not only for user preference and average performance, but for the stability and error properties required for valid inference.\footnote{See e.g., \citealt{camuffo2026varianceawarellmannotationstrategy, xu2026enhancing} for recent work in this area.}

Our results also have implications for research practice. For individual studies, model choice and configuration should be treated as part of the research design and, where feasible, specified in advance. Sensitivity analyses that rerun key estimations with multiple strong but distinct models would then play a role analogous to traditional robustness checks. Furthermore, hybrid workflows, where LLMs handle the bulk of routine annotation while human coders and design-based estimators are used to calibrate and correct systematic errors on strategically chosen subsets, may offer a balance between scalability and inferential rigor. Finally, given the variability of LLM-derived results, rigorous, detailed documentation and the establishment of reproducibility standards in LLM-assisted scientific workflow should be prioritized.

Our study has several limitations. We focus on two reasoning benchmarks and two empirical applications, and we examine a particular snapshot of model capabilities. Future work should investigate how benchmark disagreement and its downstream consequences evolve as models change, and whether similar patterns arise in domains beyond those we consider here, including tasks that go beyond text annotation. We also treat disagreement primarily as a challenge for reproducibility, but there may be settings in which epistemic diversity across models is a resource that could be strategically harnessed -- for instance, to quantify uncertainty, surface alternative pathways, or stress-test substantive conclusions. Exploring when disagreement should be suppressed, corrected, or harnessed remains an open question.

Taken together, our findings show that benchmark accuracy does not imply inferential interchangeability. LLMs enable forms of scientific analysis that were previously out of reach, and their potential to accelerate discovery is immense. At the same time, the benchmark illusion we document reveals that models with similar headline performance can diverge in ways that meaningfully affect scientific conclusions. As LLMs become integral to research workflows, recognizing and addressing this illusion will be essential for increasing reproducibility and for guiding the development of AI systems aligned with the needs of science. Doing so will require benchmarks, models, and research practices to move beyond current paradigms, so that the growing power of LLMs strengthens rather than compromises the evidentiary foundations of science.


\clearpage
\onehalfspacing
\bibliographystyle{apsr}
\bibliography{ref}

@article{barrie2024replication,
  title={Replication for language models problems, principles, and best practice for political science},
  author={Barrie, Christopher and Palmer, Alexis and Spirling, Arthur},
  year={2024}
}

@article{gelman2013garden,
  title={The garden of forking paths: Why multiple comparisons can be a problem, even when there is no “fishing expedition” or “p-hacking” and the research hypothesis was posited ahead of time},
  author={Gelman, Andrew and Loken, Eric},
  journal={Department of Statistics, Columbia University},
  volume={348},
  number={1-17},
  pages={3},
  year={2013}
}

@article{wang2024mmlu,
  title={Mmlu-pro: A more robust and challenging multi-task language understanding benchmark},
  author={Wang, Yubo and Ma, Xueguang and Zhang, Ge and Ni, Yuansheng and Chandra, Abhranil and Guo, Shiguang and Ren, Weiming and Arulraj, Aaran and He, Xuan and Jiang, Ziyan and others},
  journal={Advances in Neural Information Processing Systems},
  volume={37},
  pages={95266--95290},
  year={2024}
}

@inproceedings{rein2024gpqa,
  title={Gpqa: A graduate-level google-proof q\&a benchmark},
  author={Rein, David and Hou, Betty Li and Stickland, Asa Cooper and Petty, Jackson and Pang, Richard Yuanzhe and Dirani, Julien and Michael, Julian and Bowman, Samuel R},
  booktitle={First Conference on Language Modeling},
  year={2024}
}

@article{liang2022holistic,
  title={Holistic evaluation of language models},
  author={Liang, Percy and Bommasani, Rishi and Lee, Tony and Tsipras, Dimitris and Soylu, Dilara and Yasunaga, Michihiro and Zhang, Yian and Narayanan, Deepak and Wu, Yuhuai and Kumar, Ananya and others},
  journal={arXiv preprint arXiv:2211.09110},
  year={2022}
}

@article{egami2024using,
  title={Using large language model annotations for the social sciences: A general framework of using predicted variables in downstream analyses},
  author={Egami, Naoki and Hinck, Musashi and Stewart, Brandon M and Wei, Hanying},
  journal={Preprint from November},
  volume={17},
  pages={2024},
  year={2024}
}

@article{kim2021improving,
  title={Improving elementary grade students’ science and social studies vocabulary knowledge depth, reading comprehension, and argumentative writing: A conceptual replication},
  author={Kim, James S and Relyea, Jackie Eunjung and Burkhauser, Mary A and Scherer, Ethan and Rich, Patrick},
  journal={Educational Psychology Review},
  volume={33},
  number={4},
  pages={1935--1964},
  year={2021},
  publisher={Springer}
}

@article{rozenas2019autocrats,
  title={How autocrats manipulate economic news: Evidence from Russia’s state-controlled television},
  author={Rozenas, Arturas and Stukal, Denis},
  journal={The Journal of Politics},
  volume={81},
  number={3},
  pages={982--996},
  year={2019},
  publisher={The University of Chicago Press Chicago, IL}
}

@article{baumann2025large,
  title={Large Language Model Hacking: Quantifying the Hidden Risks of Using LLMs for Text Annotation},
  author={Baumann, Joachim and R{\"o}ttger, Paul and Urman, Aleksandra and Wendsj{\"o}, Albert and Plaza-del-Arco, Flor Miriam and Gruber, Johannes B and Hovy, Dirk},
  journal={arXiv preprint arXiv:2509.08825},
  year={2025}
}

@article{ziems2024can,
  title={Can large language models transform computational social science?},
  author={Ziems, Caleb and Held, William and Shaikh, Omar and Chen, Jiaao and Zhang, Zhehao and Yang, Diyi},
  journal={Computational Linguistics},
  volume={50},
  number={1},
  pages={237--291},
  year={2024},
  publisher={MIT Press One Broadway, 12th Floor, Cambridge, Massachusetts 02142, USA~…}
}

@article{angelopoulos2023prediction,
  title={Prediction-powered inference},
  author={Angelopoulos, Anastasios N and Bates, Stephen and Fannjiang, Clara and Jordan, Michael I and Zrnic, Tijana},
  journal={Science},
  volume={382},
  number={6671},
  pages={669--674},
  year={2023},
  publisher={American Association for the Advancement of Science}
}

@misc{gil2025accelerating,
  title={Accelerating science with AI},
  author={Gil, Dar{\'\i}o and Moler, Kathryn A},
  journal={Science},
  pages={eaee0605},
  year={2025},
  publisher={American Association for the Advancement of Science}
}

@article{zhang2025exploring,
  title={Exploring the role of large language models in the scientific method: from hypothesis to discovery},
  author={Zhang, Yanbo and Khan, Sumeer A and Mahmud, Adnan and Yang, Huck and Lavin, Alexander and Levin, Michael and Frey, Jeremy and Dunnmon, Jared and Evans, James and Bundy, Alan and others},
  journal={npj Artificial Intelligence},
  volume={1},
  number={1},
  pages={14},
  year={2025},
  publisher={Nature Publishing Group UK London}
}

@article{cui2025large,
  title={A large-scale replication of scenario-based experiments in psychology and management using large language models},
  author={Cui, Ziyan and Li, Ning and Zhou, Huaikang},
  journal={Nature Computational Science},
  volume={5},
  number={8},
  pages={627--634},
  year={2025},
  publisher={Nature Publishing Group US New York}
}

@article{argyle2023out,
  title={Out of one, many: Using language models to simulate human samples},
  author={Argyle, Lisa P and Busby, Ethan C and Fulda, Nancy and Gubler, Joshua R and Rytting, Christopher and Wingate, David},
  journal={Political Analysis},
  volume={31},
  number={3},
  pages={337--351},
  year={2023},
  publisher={Cambridge University Press}
}

@unpublished{bisbee2025human,
  author = {Bisbee, J. and Spirling, A.},
  title = {What to Do When Humans Are No Longer the Gold Standard: Large Language Models, State of the Art and Robustness},
  year = {2025}
}

@article{gao2024quantifying,
  title={Quantifying the use and potential benefits of artificial intelligence in scientific research},
  author={Gao, Jian and Wang, Dashun},
  journal={Nature human behaviour},
  volume={8},
  number={12},
  pages={2281--2292},
  year={2024},
  publisher={Nature Publishing Group UK London}
}

@book{krippendorff2018content,
  title={Content analysis: An introduction to its methodology},
  author={Krippendorff, Klaus},
  year={2018},
  publisher={Sage publications}
}

@article{hao2024ai,
  title={AI Expands Scientists' Impact but Contracts Science's Focus},
  author={Hao, Qianyue and Xu, Fengli and Li, Yong and Evans, James},
  journal={arXiv preprint arXiv:2412.07727},
  year={2024}
}

@article{westwood2025potential,
  title={The potential existential threat of large language models to online survey research},
  author={Westwood, Sean J},
  journal={Proceedings of the National Academy of Sciences},
  volume={122},
  number={47},
  pages={e2518075122},
  year={2025},
  publisher={National Academy of Sciences}
}

@article{kusumegi2025scientific,
  title={Scientific production in the era of large language models},
  author={Kusumegi, Keigo and Yang, Xinyu and Ginsparg, Paul and de Vaan, Mathijs and Stuart, Toby and Yin, Yian},
  journal={Science},
  volume={390},
  number={6779},
  pages={1240--1243},
  year={2025},
  publisher={American Association for the Advancement of Science}
}

@article{gilardi2023chatgpt,
  title={ChatGPT outperforms crowd workers for text-annotation tasks},
  author={Gilardi, Fabrizio and Alizadeh, Meysam and Kubli, Ma{\"e}l},
  journal={Proceedings of the National Academy of Sciences},
  volume={120},
  number={30},
  pages={e2305016120},
  year={2023},
  publisher={National Academy of Sciences}
}

@article{agrawal2022large,
  title={Large language models are few-shot clinical information extractors},
  author={Agrawal, Monica and Hegselmann, Stefan and Lang, Hunter and Kim, Yoon and Sontag, David},
  journal={arXiv preprint arXiv:2205.12689},
  year={2022}
}

@article{d2022underspecification,
  title={Underspecification presents challenges for credibility in modern machine learning},
  author={D'Amour, Alexander and Heller, Katherine and Moldovan, Dan and Adlam, Ben and Alipanahi, Babak and Beutel, Alex and Chen, Christina and Deaton, Jonathan and Eisenstein, Jacob and Hoffman, Matthew D and others},
  journal={Journal of Machine Learning Research},
  volume={23},
  number={226},
  pages={1--61},
  year={2022}
}

@article{reiss2023testing,
  title={Testing the reliability of chatgpt for text annotation and classification: A cautionary remark},
  author={Reiss, Michael V},
  journal={arXiv preprint arXiv:2304.11085},
  year={2023}
}

@article{messeri2024artificial,
  title={Artificial intelligence and illusions of understanding in scientific research},
  author={Messeri, Lisa and Crockett, Molly J},
  journal={Nature},
  volume={627},
  number={8002},
  pages={49--58},
  year={2024},
  publisher={Nature Publishing Group UK London}
}

@article{carlson2025use,
  title={The use of LLMs to annotate data in management research: Foundational guidelines and warnings},
  author={Carlson, Natalie A and Burbano, Vanessa},
  journal={Strategic Management Journal},
  year={2025},
  publisher={Wiley Online Library}
}

@article{xu2026enhancing,
  title={Enhancing LLM-Based Data Annotation with Error Decomposition},
  author={Xu, Zhen and Khatri, Vedant and Dai, Yijun and Liu, Xiner and Li, Siyan and Zhang, Xuanming and Yu, Renzhe},
  journal={arXiv preprint arXiv:2601.11920},
  year={2026}
}

@misc{camuffo2026varianceawarellmannotationstrategy,
      title={Variance-Aware LLM Annotation for Strategy Research: Sources, Diagnostics, and a Protocol for Reliable Measurement}, 
      author={Arnaldo Camuffo and Alfonso Gambardella and Saeid Kazemi and Jakub Malachowski and Abhinav Pandey},
      year={2026},
      eprint={2601.02370},
      archivePrefix={arXiv},
      primaryClass={cs.CY}
}

\pagebreak
\appendix
\onehalfspacing
\setcounter{page}{1}
\setcounter{table}{0}
\setcounter{figure}{0}
\setcounter{equation}{0}
\setcounter{footnote}{0}
\renewcommand\thetable{A\arabic{table}}
\renewcommand\thefigure{A\arabic{figure}}
\renewcommand{\thepage}{A-\arabic{page}}
\renewcommand{\theequation}{A\arabic{equation}}
\renewcommand{\thefootnote}{A\arabic{footnote}}

\noindent \textbf{\large Appendix \textit{for}}
\bigskip\begin{center}
    \textbf{\large Benchmark Illusion: Disagreement among LLMs and Its Scientific Consequences}\\
    \bigskip
\end{center}

\vspace{1em}
\noindent\hspace{0em}{\large\bf\underline{Table of Contents}}

	\begin{enumerate}\itemsep0ex\small
        \bf\item[A.] Additional annotation details
		\bf\item[B.] Additional result: \citet{kim2021improving}
            \bf\item[C.] Additional result: \citet{rozenas2019autocrats}
	\end{enumerate}
    
\thispagestyle{empty} 
\clearpage
\setcounter{page}{1}

\doublespacing

\section{Additional annotation details}\label{annotation}

We download all LLMs used in the study from Hugging Face (\url{https://huggingface.co/}). We use \texttt{vllm} as the inference engine for all annotations. To ensure reproducibility, we follow the reproducibility guide from \texttt{vllm} (\url{https://docs.vllm.ai/en/v0.12.0/examples/offline_inference/reproducibility/?h=reproducibility}) by setting the environment variable \say{VLLM\_ENABLE\_V1\_MULTIPROCESSING} to 0. Additionally, we use greedy decoding by setting the temperature to zero. We use Nvidia H100 GPUs for all model inference.

\section{Additional result: \citet{kim2021improving}}\label{kim}

\begin{figure}[h]
\begin{center}
\caption{Reanalysis of \citet{kim2021improving}}
\label{fig:education_sci}
\includegraphics[scale=0.5]{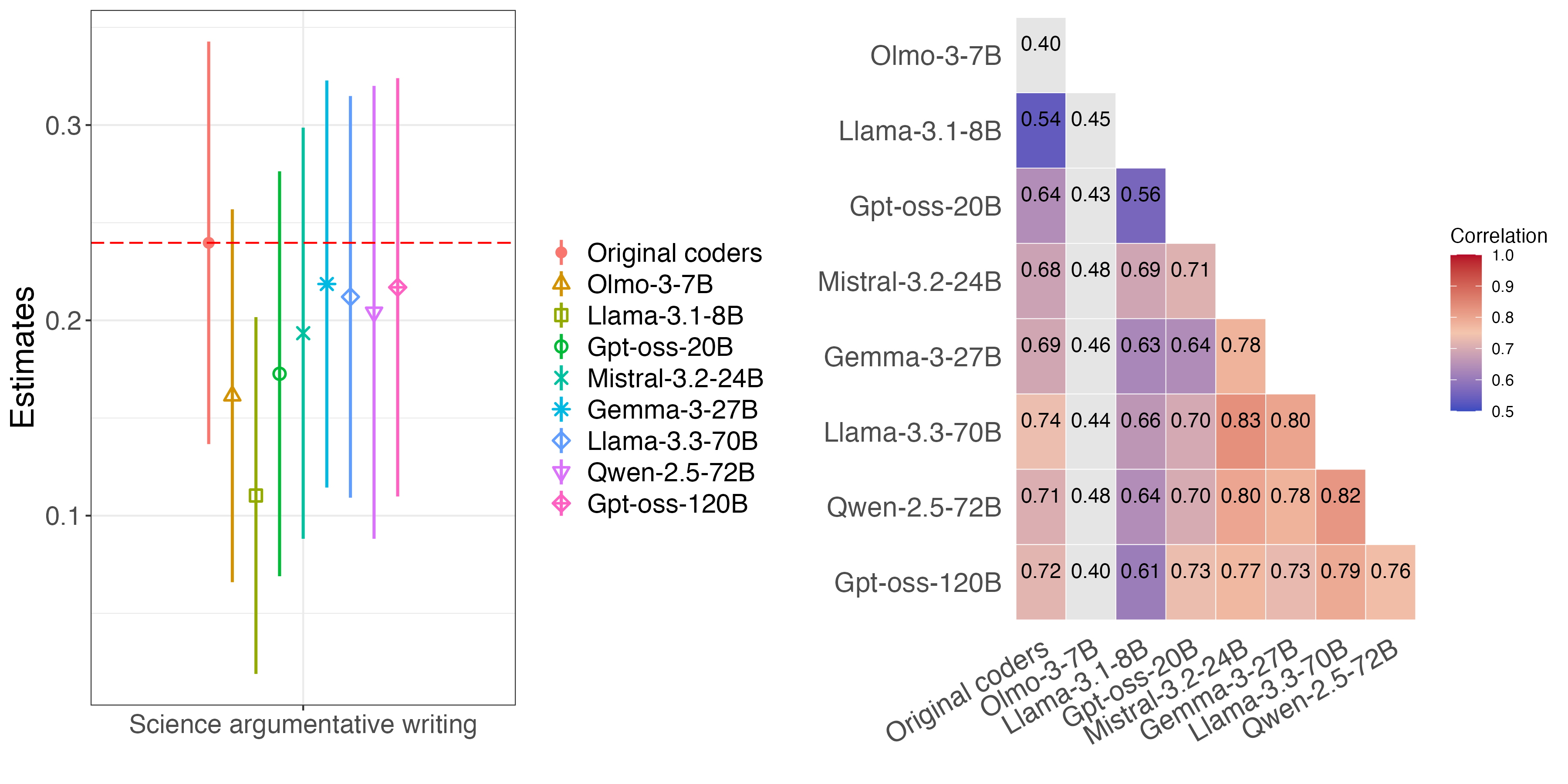}
\end{center}
\vspace{-2ex}
{\footnotesize \textbf{Notes:} The left panel shows the estimated treatment effect of a literacy intervention on student argumentative writing scores. Each point represents an estimate using scores from a different annotator (LLMs or the original human-coded study data), with bars indicating 95\% confidence intervals. The red dashed line marks the original study's point estimate. The right panel shows the correlation of the scores by a given pair of coders.}
\end{figure}

Figure \ref{fig:education_sci} presents the reanalysis results on the estimated treatment effect on science argumentative writing for the original study (based on human coders) and for each of the eight LLMs. The right panel shows the correlations of the scores by a given pair of coders.

Similar to the results on social studies argumentative writing, Figure \ref{fig:education_sci} shows that the magnitude of the estimated effect varies substantially depending on the model used for annotation, ranging from a low of 0.11 for Llama-3.1-8B to a high of 0.22 for Gemma-3-27B. 

\pagebreak

\section{Additional result: \citet{rozenas2019autocrats}}\label{rozenas}

\begin{figure}[h]
\begin{center}
\caption{Reanalysis of \citet{rozenas2019autocrats}}
\label{fig:polsci_putin}
\includegraphics[scale=0.5]{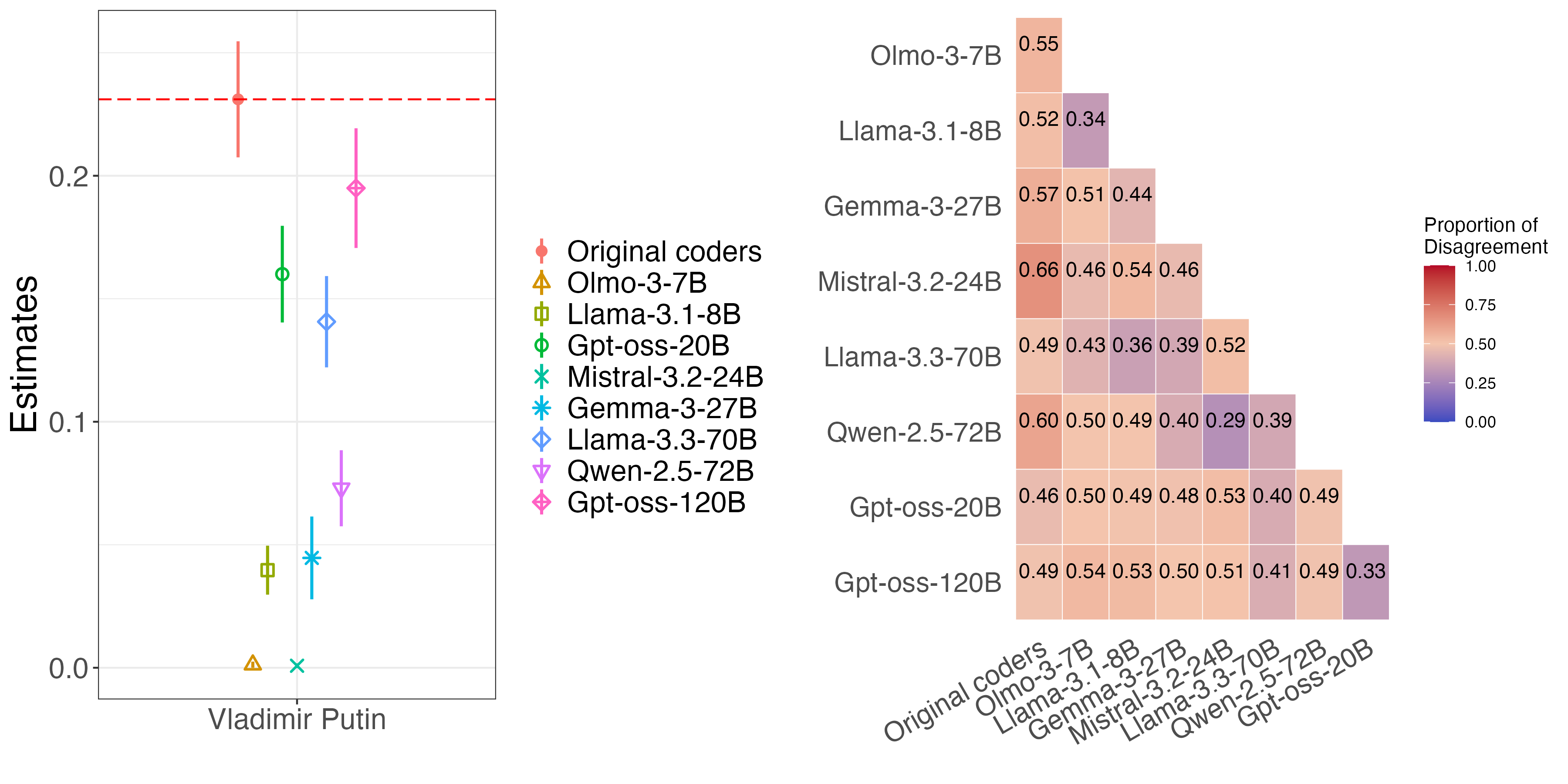}
\end{center}
\vspace{-2ex}
{\footnotesize \textbf{Notes:} The left panel shows the estimated relative attribution of economic news to \say{Vladimir Putin.} Each point is an estimate using annotations from a different source (LLMs or the original crowdsourced data), with bars indicating 95\% confidence intervals. The red dashed line marks the original study's point estimate. A positive value indicates that good news is more likely to be attributed to Vladimir Putin than bad news. The right panel shows the pairwise proportion of disagreement among the different annotators.}
\end{figure}

\begin{figure}[h]
\begin{center}
\caption{Reanalysis of \citet{rozenas2019autocrats}}
\label{fig:polsci_forec}
\includegraphics[scale=0.5]{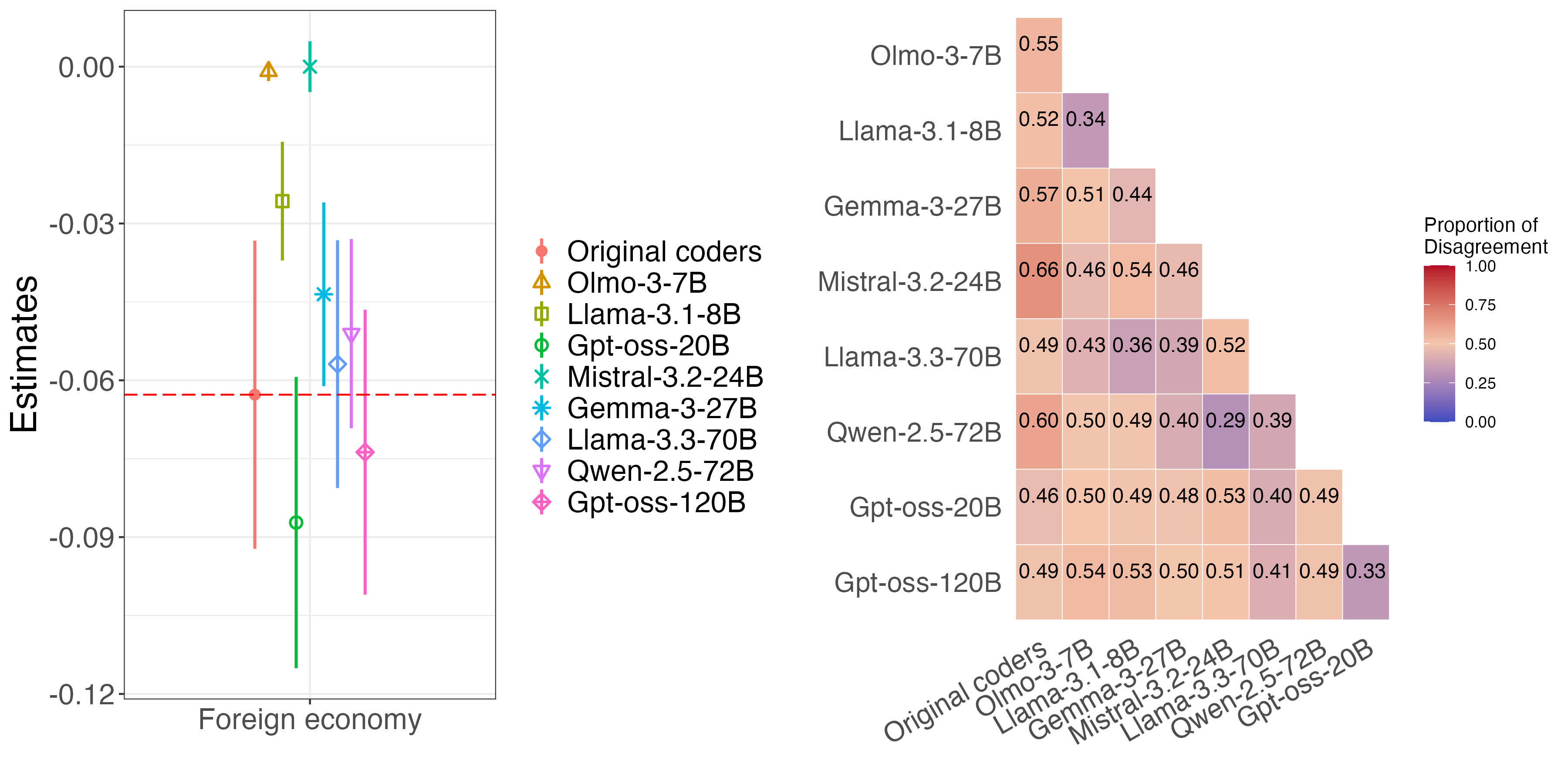}
\end{center}
\vspace{-2ex}
{\footnotesize \textbf{Notes:} The left panel shows the estimated relative attribution of economic news to \say{foreign economies.} Each point is an estimate using annotations from a different source (LLMs or the original crowdsourced data), with bars indicating 95\% confidence intervals. The red dashed line marks the original study's point estimate. A positive value indicates that good news is more likely to be attributed to foreign economies than bad news. The right panel shows the pairwise proportion of disagreement among the different annotators.}
\end{figure}

\begin{figure}[h]
\begin{center}
\caption{Reanalysis of \citet{rozenas2019autocrats}}
\label{fig:polsci_forgov}
\includegraphics[scale=0.5]{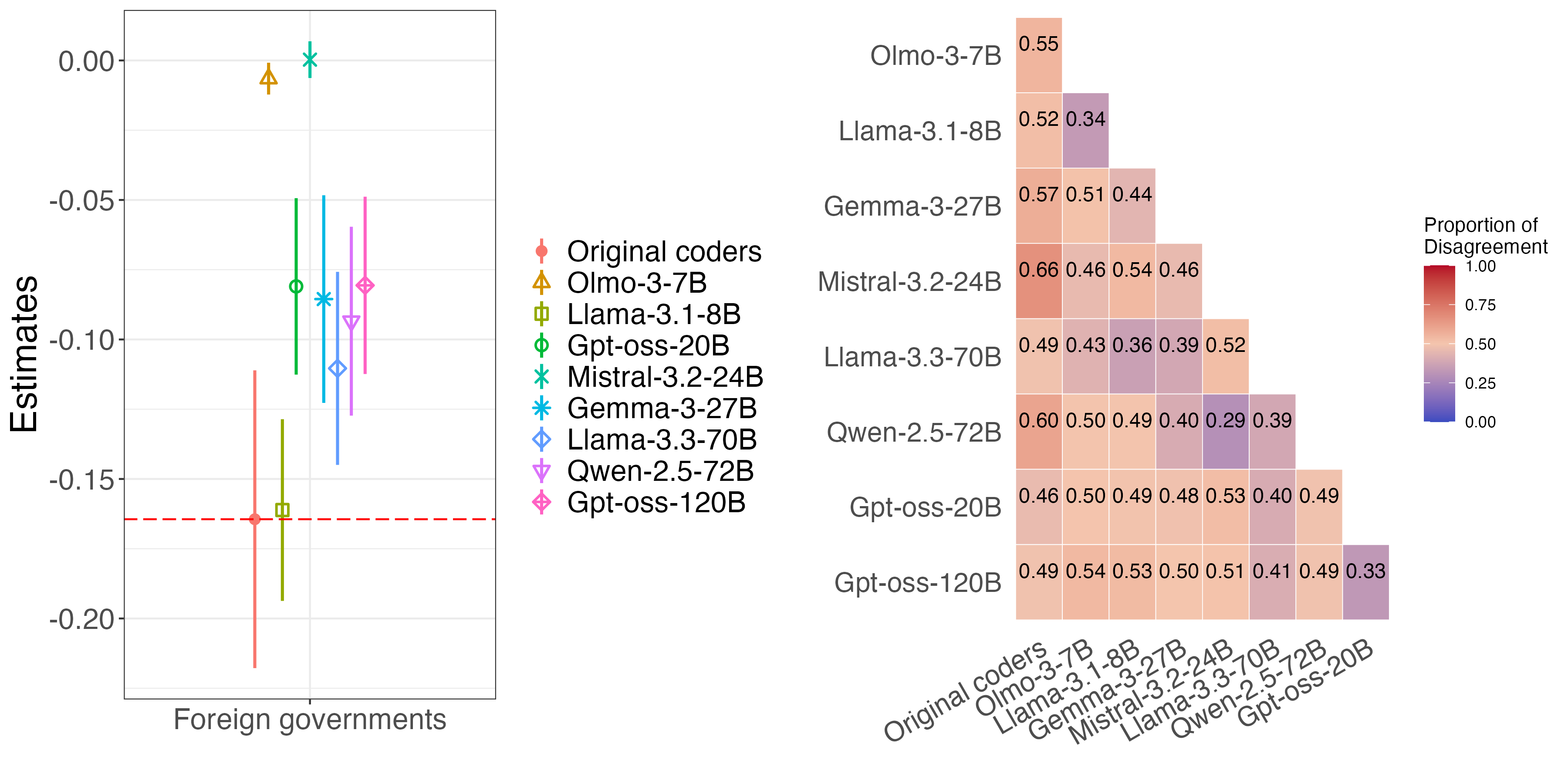}
\end{center}
\vspace{-2ex}
{\footnotesize \textbf{Notes:} The left panel shows the estimated relative attribution of economic news to \say{foreign governments.} Each point is an estimate using annotations from a different source (LLMs or the original crowdsourced data), with bars indicating 95\% confidence intervals. The red dashed line marks the original study's point estimate. A positive value indicates that good news is more likely to be attributed to foreign governments than bad news. The right panel shows the pairwise proportion of disagreement among the different annotators.}
\end{figure}

Figures \ref{fig:polsci_putin}, \ref{fig:polsci_forec}, and \ref{fig:polsci_forgov} present the reanalysis results on the selective attribution of the other three entities in the original paper: Vladimir Putin, foreign economies, and foreign governments. Similar to the result for Russian officials, all outcomes show a large variation in coefficient estimates.

\end{document}